\documentclass[lettersize,journal]{IEEEtran}
\usepackage{amsmath,amsfonts}
\usepackage{algorithmic}
\usepackage{algorithm}
\usepackage{array}
\usepackage[caption=false,font=normalsize,labelfont=sf,textfont=sf]{subfig}
\usepackage{textcomp}
\usepackage{stfloats}
\usepackage{url}
\usepackage{verbatim}
\usepackage{graphicx}
\usepackage{cite}
\usepackage{chemfig}
\usepackage{xcolor}

\begin{document}

\title{UMDFood: Vision-Language Models Boost Food Composition Compilation}

\author{Peihua Ma$^*$, Yixin Wu$^*$, Ning Yu, Yang Zhang, Michael Backes, Qin Wang, Cheng-I Wei \thanks{$^*$ Equal contribution.}}

\maketitle

\begin{abstract}
Nutrition information is crucial in precision nutrition and the food industry. The current food composition compilation paradigm relies on laborious and experience-dependent methods. However, these methods struggle to keep up with the dynamic consumer market, resulting in delayed and incomplete nutrition data. In addition, earlier machine learning methods overlook the information in food ingredient statements or ignore the features of food images. To this end, we propose a novel vision-language model, UMDFood-VL, using front-of-package labeling and product images to accurately estimate food composition profiles. In order to empower model training, we established UMDFood-90k, the most comprehensive multimodal food database to date, containing 89,533 samples, each labeled with image and text-based ingredient descriptions and 11 nutrient annotations. UMDFood-VL achieves the macro-AUCROC up to 0.921 for fat content estimation, which is significantly higher than existing baseline methods and satisfies the practical requirements of food composition compilation. Meanwhile, up to 82.2\% of selected products' estimated error between chemical analysis results and model estimation results are less than 10\%. This performance sheds light on generalization towards other food and nutrition-related data compilation and catalyzation for the evolution of generative AI-based technology in other food applications that require personalization.
\end{abstract}


\section{Introduction}
\IEEEPARstart{P}{recision} nutrition, also referred to as personalized nutrition, places an emphasis on the individual as opposed to a group of people\cite{berry2020human}. Precision nutrition is specific to develop interventions to prevent or treat chronic diseases based on race, gender, health history, DNA, and lifestyle habits\cite{martinez2021personalised}. Accurate nutrition data is the pillar of precision nutrition which is collected by food composition compilation\cite{san2020contribution,liu2022future}. Compilation of food composition data refers to the systematic collection, organization, and presentation of information related to the nutritional components found in different foods\cite{rand1991compiling,greenfield2003food}. 

The food composition data compiling method has not fundamentally improved since the first food composition tables of Wilbur O Atwater in 1894 to the up-to-date USDA FoodData Central\cite{rand1991compiling}. Currently, food composition compilation is achieved by the hybrid method, which combines the direct method collected data by chemical analysis and indirect inherited data which are estimates derived from analytical values for a comparable food product\cite{kapsokefalou2019food}. However, limited by labor and resources, food composition databases do not collect products that are less common or are traditional to certain cultures\cite{ene2019importance}. Meanwhile, the database covers few nutrients or bioactive compounds present in foods\cite{parodi2018potential}. In addition, nutrient information can become outdated as farming methods, food fortification policies, and analytical techniques evolve\cite{delgado2021food}. To overcome these drawbacks, machine learning methods including linear SVM\cite{barbosa2016recognition}, Bayesian\cite{van2023moisture}, MLP\cite{ma2021application}, and CNN\cite{ma2022application} were applied to predict composition profiles based on known information, e.g. ingredients, image, and genetic information of the products. However, these methods cannot satisfy the accuracy and precision requirements for large-scale food composition compilation due to the high diversity and complexity of the task.

Traditional machine learning models typically operate in isolated domains, either focusing on image recognition or text analysis\cite{ray2019quick}. In contrast, we introduce a vision-language model, UMDFood-VL, which integrates these two types of data, enabling a much richer contextual understanding from the visual cues and textual information of food items. For instance, the model could analyze both the picture and the ingredient statement of a food item to predict its nutritional composition, capturing nuances that might be missed when considering only one type of data. Meanwhile, traditional machine learning models often struggle with unstructured or semi-structured data. The vision-language model is designed to tackle this complexity head-on. It can process untagged photos from search engines, or unstructured oral descriptions of food products, converting them into actionable insights about nutritional composition. 

Current food composition databases have limitations that prevent them from providing complete and reliable data\cite{kirk2021precision}. While these databases have become more sophisticated over time, nutrient facts can still be overgeneralized (e.g. neglecting origin, vintage impact on nutrient content), and functional nutrient data is often missing (e.g. phytochemicals in products). Moreover, the large number of food products on the market makes it impractical to analyze the composition of every single item using traditional methods of compilation. As a result, it is challenging to obtain accurate and comprehensive information about the nutritional content of food\cite{rand1991compiling}. Improvements in data collection, as well as analysis methods, are needed to address these issues and provide more reliable nutrition information for precision health\cite{menichetti2023machine}. As a result, we introduce our new database, UMDFood-90k, which is the largest food database to the best of our knowledge, which covers 89,533 products inherited from the USDA branded food product database (USDA-BFPD), and exponentially increasing the size of the current food database.  Based on our novel model and new dataset, we achieve 0.921 of macro-AUCROC for fat content estimation, outperforming the recent baseline method by Jiang et al\cite{min2023large}, and Ma et al\cite{ma2021application}. The average AUCROC for all selected nutrients is 0.827, significantly higher than any other previous report models. Further validation shows that 82.2\% of our nutrient estimation makes less than 10\% errors from the chemical analysis ground truth, which is recognized as acceptable for practice requirement.

\begin{figure*}[!t]
\centering
\includegraphics[width=7in]{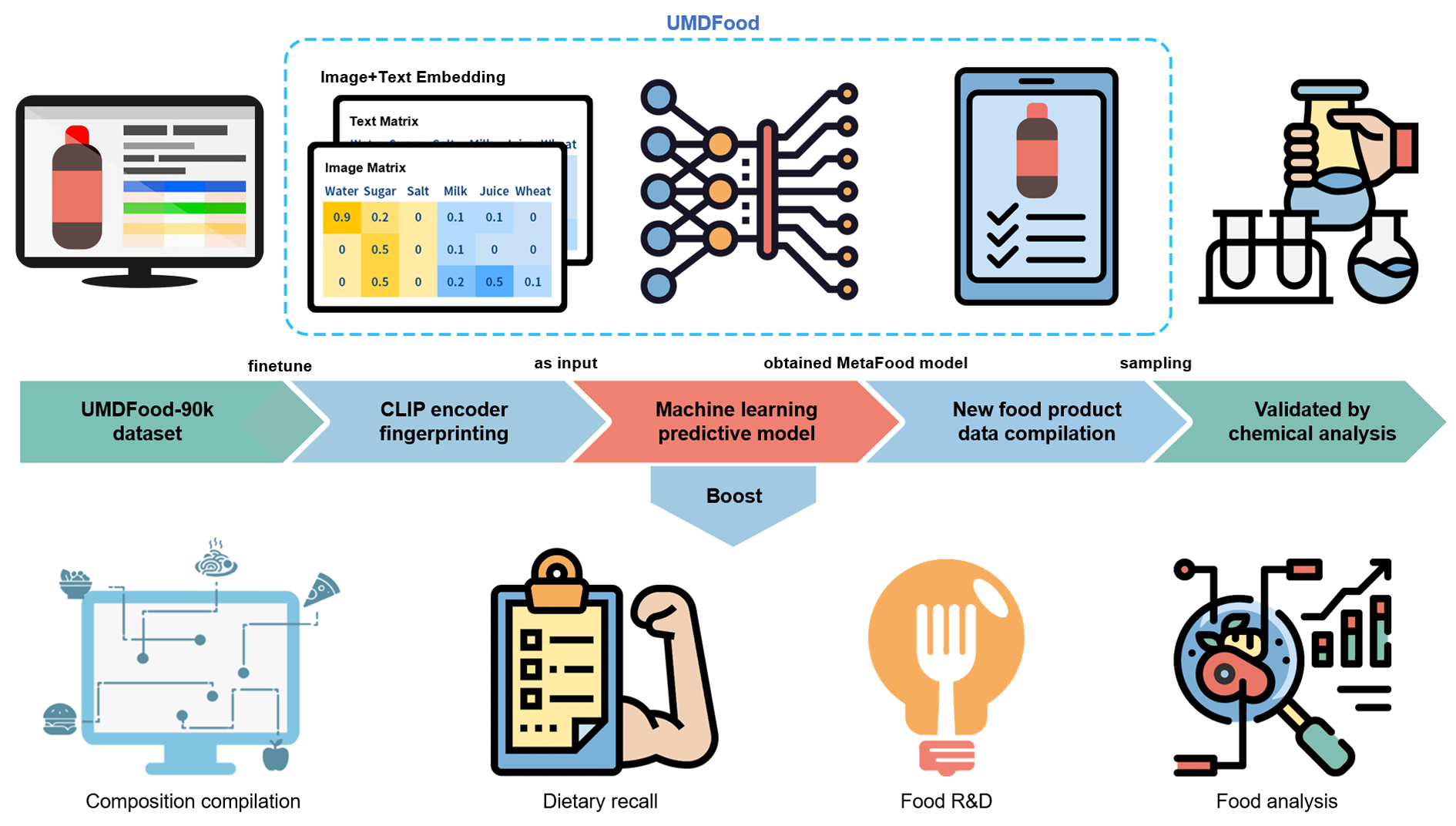}
\caption{The conceptual pipeline of our research. Our contribution includes collecting the UMDFood-90k multimodal dataset and developing UMDFood models for food composition compilation which may boost applications in precision nutrition, food R\&D, and food analysis. }
\label{fig_1}
\end{figure*}

Our contributions are listed in four thrusts.
\begin{itemize}
    \item We pioneer to application of vision-language models, UMDFood-VL, for food composition compilation, excelling at fusing textual and visual information of food items into a unified representation and capturing more nutrient information than ever before. 
    \item To empower model training, we collect the UMDFood-90k dataset, which is the largest food multi-media dataset up-to-date, containing 89,533 samples, each associated with a product image, text-based ingredient statement, and six nutrient amount annotations.
    \item We develop the UMDFood-VL model (fig.~\ref{fig_1}). The UMDFood-VL model is finetuned with the dataset to generate embeddings for food images and ingredients, improving the accuracy of all nutrient estimations. UMDFood-VL achieves up to 92.2\% macro-AUCROC for fat content.
    \item We validate the model result by chemical analysis. Up to 82.2\% of validated samples' difference between chemical analysis data and model estimate results is less than 10\%.
\end{itemize}

\section{Related work}
\subsection{Food composition compilation}
The compilation method was distinguished by Southgate as the direct and indirect method of compiling tables in 1974\cite{greenfield2003food}. The direct method analyzed different purchases of the same food separately to easily yield exceptionally reliable data. But only limited information on nutrient variation in each food is obtained and this method imposes pressure on the analytical resources. The indirect method uses data collected from published literature (e.g., meta-analysis) or unpublished laboratory reports (e.g., USDA standard release)\cite{pennington2008applications}. Evidently, quality appraisal for inclusion in the database by the indirect method is inevitable since original data sources are consequently less controlled and are therefore of a lower degree of confidence\cite{pennington2008applications}. Nowadays, most food compilation workers take advantage of both the direct and indirect methods by combining the original analytical values, directly borrowed values, and the calculated values which are called indirectly inherited values together\cite{greenfield2003food}. These indirect inherited values include estimates derived from analytical values for a comparable food (e.g. values for fuji apple used for gala apple)\cite{guyot1998reversed}, for another form of the same item (e.g. values for the “roasted” food used for “boiled” food)\cite{wandsnider1997roasted}, calculation from incomplete analyses of samples (e.g. carbohydrate by difference)\cite{menezes2004measurement} and similar imputation by comparing data for different forms of the same food (e.g. “fresh” versus “defatted”)\cite{zhu2006proteins} as well as derived from nutrient content of the ingredients and corrected for retention factors (e.g. recipes vary dramatically, using specific algorithms, retention and yield factors)\cite{charrondiere2011report}. Hitherto, these calculations and compilations solely rely on the expertise of database managers and are costly, time-consuming, vulnerable, and lagging updates\cite{marconi2018food}. 
Machine learning and deep learning offer potential solutions to the limitations of traditional methods for obtaining comprehensive nutrition information\cite{min2023large}. However, previous methods based on food image recognition have proven impractical due to the limited size of their datasets (previous reports dataset only covered 2,000 classes of food, which is still insufficient for precision nutrition practice) and the complexity of certain food items, such as salads with multiple ingredients that cannot be recognized accurately\cite{min2023large,wang2022review}. 

\subsection{Vision-language models}
As multimedia data continues to proliferate from various sources, a wide array of downstream tasks have emerged. 
The downstream tasks can be mainly categorized into 4 categories~\cite{chen2023vlp}: the classification tasks (e.g., visual question answering and natural language for visual reasoning), the regression tasks (e.g., multi-modal sentiment analysis), the retrieval tasks (vision-language retrieval), and the generation tasks (e.g., visual captioning and visual dialogue). 
All these tasks require the collaborative integration of both visual and linguistic knowledge and have attracted massive attention in recent years.
In response to this rapidly growing demand, a substantial number of vision-language models (e.g., CLIP~\cite{radford2021learning}, SLIP~\cite{mu2022slip}, ALIGN~\cite{jia2021scaling}, BLIP~\cite{li2022blip}, BLIP-2~\cite{li2023blip}, Flamingo~\cite{alayrac2022flamingo}, MiniGPT-4~\cite{zhu2023minigpt}, and InstructBLIP~\cite{liu2023improved}) designed to bridge the gap between visual and linguistic modalities, have been proposed and have made significant advancements.
In the field of nutrition, text-based training relying purely on ingredient lists has faced challenges due to over-sparse and biased ingredient distributions\cite{ma2022deep}.

Considering the recent remarkable achievements of vision-language models, it is intuitive to build a bridge between these models and practical applications to unleash the full potential of these cutting-edge technologies on food nutrition estimation tasks.
Hence, we propose the UMDFood model, i.e., a vision-language model, that has the ability to connect the information from the vision modality (the product image) and from the language modality (the ingredient statement).
Different from the traditional food composition compilation method, our method takes advantage of the vision-language model and achieves high accuracy on large-scale food products.
Specifically, we leverage CLIP, one of the most representative vision-language models, to construct the UMDFood model.

The CLIP model has an image encoder $E_I$ and a text encoder $E_T$ to connect the image data with corresponding text descriptions in a common latent space. 
Hence, the CLIP model is usually pre-trained on large-scale image-text pairs $D = \{(\mathbf{I}_i, \mathbf{T}_i)\}_{i=1}^N$.
The CLIP model employs the InfoNCE loss~\cite{OLV18} to optimize both $E_I$ and $E_T$, ensuring that the embeddings from the same pair have high cosine similarity while those from different pairs have low similarity.
In a batch of $N$ image-text pairs, the loss for the image encoder $E_I$ is formulated as follows:
\begin{equation}
       L_I = -\frac{1}{N}\sum_{i=1}^{N}\log\frac{\exp\Big(\mathsf{sim}\big(E_I(\mathbf{I}_i), E_T(\mathbf{T}_i)\big)/\tau\Big)}{\sum^N_{j=1}\exp\Big(\mathsf{sim}\big(E_I(\mathbf{I}_i), E_T(\mathbf{T}_j)\big)/\tau\Big)}
\end{equation}
where $\mathsf{sim}(\cdot, \cdot)$ represents the cosine similarity between two features, and $\tau$ serves as a temperature parameter regulating the similarity distribution.
The CLIP model has a symmetrical loss for the image encoder $E_I$ and the text encoder $E_T$, so the loss for $E_T$ is formulated as follows:
\begin{equation}
    L_T = -\frac{1}{N}\sum_{i=1}^{N}\log\frac{\exp\Big(\big(E_T(\mathbf{T}_i), E_I(\mathbf{I}_i)\big)/\tau\Big)}{\sum^N_{j=1}\exp\Big(\mathsf{sim}\big(E_T(\mathbf{T}_i), E_I(\mathbf{I}_j)\big)/\tau\Big)}
\end{equation}
Overall, the loss function $L_\text{CLIP}$ for training/fine-tuning the CLIP model is formally defined as follows.
\begin{equation}
    L_\text{CLIP} = (L_I + L_T) / 2
\end{equation}
In this paper, we leverage the publicly available CLIP pre-trained weights as a good initialization to train our UMDFood model.

Text-based training relying purely on ingredient lists has also faced challenges due to over-sparse and biased ingredient distributions\cite{ma2022deep}. Recent breakthroughs in transformer models have provided innovative and automated solutions to these labor-intensive processes.  The milestone works including self-supervision meets contrastive language-image pre-training (CLIP) models~\cite{radford2021learning}, language-image pre-training (SLIP)~\cite{mu2022slip}, bootstrapping language-image pre-training (BLIP)\cite{li2022blip}, large-scale image and noisy-text embedding pre-training (ALIGN)\cite{jia2021scaling}, and object-semantics aligned pre-training (OSCAR)\cite{vedaldi2020computer}. However, despite these advancements, significant challenges remain in applying these technologies to specific scenarios due to professional barriers and technical limitations. Building a bridge between vision-language models and practical applications will be crucial to realizing the full potential of these cutting-edge technologies in the field of nutrition. Different from the traditional food composition compilation method, our method takes advantage of the vision-language model and achieves high accuracy on large-scale food products. 
\section{Our approach}
\label{section:approach}

\begin{figure*}[!t]
\centering
\includegraphics[width=\linewidth]{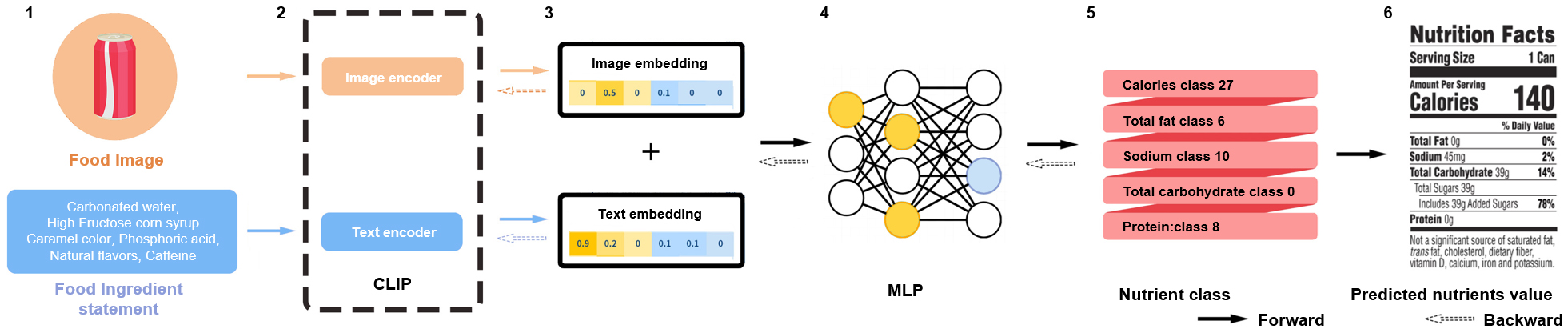}
\caption{The UMDFood-VL model consists of a CLIP model and an MLP classifier. The CLIP model takes product images and ingredient statements as input and generates image embeddings and text embeddings to feed into the following 3-layer MLP classifier. During training, we first initialized the CLIP model with the pre-trained weights from Open AI’s CLIP model. Then we optimized the CLIP model and MLP classifier simultaneously using the cross-entropy loss on the UMDFood-90k dataset.}
\label{fig_2}
\end{figure*}

\subsection{Model design details}
We developed a new vision-language AI model, named UMDFood-VL (fig.~\ref{fig_2}), for our nutrient value estimation.
More specifically, we construct the UMDFood model on top of the CLIP model, along with an MLP classifier.
Hence, the input of the UMDFood model is a pair of the product image and the corresponding ingredient statement. 
We consider the food nutrient estimation task as a classification task, and the output of the UMDFood model is the nutrient value.
Hence, to train the final model, we leverage the conventional loss for classification tasks, i.e., the cross-entropy loss, which is formally defined as follows:
\begin{equation}
    \mathcal{L} = -\frac{1}{N}\sum_{i=1}^{N}\sum^{M}_{c=1}{\hat{\mathbf{y}_i^c}}\log{F\Big(\mathsf{concat}\big(E_I(\mathbf{I}_i), E_T(\mathbf{T}_i)\big)\Big)^c}
\end{equation}
where $M$ is the number of classes, $\hat{\mathbf{y}_i^c}$ is the ground-truth binary indicator of the class $c$, $F(\cdot)^c$ is the MLP classifier's confidence score of class $c$.
Note that, during the training process, the MLP classifier $F$, the image encoder $E_I$, and the text encoder $E_T$ are optimized.

For the image encoder $E_I$, we used the Vision Transformer which has a 32$\times$32 input patch size, 12 attention layers, and 12 attention heads. The text encoder $E_T$ is a Transformer with 12 attention layers and 8 attention heads. 
The MLP classifier $F$ takes image embeddings, text embeddings, or both of them as input and the ground truth nutrient value as output. The MLP classifier contains 3 layers, and the number of neurons in each layer is set to 64, 16, and the number of classes for the given nutrient.  

\begin{figure}[!t]
\centering
\includegraphics[width=3.5in]{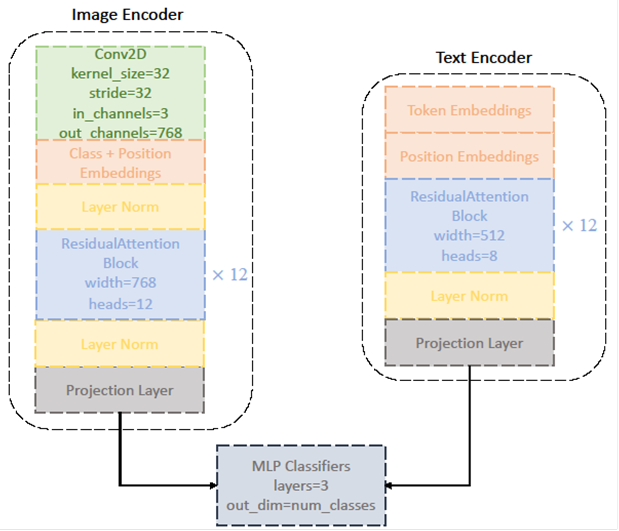}
\caption{The details of the UMDFood model architecture. }
\label{fig_3}
\end{figure}

Our classification model comprises an image encoder, a text encoder, and an MLP classifier (Fig.\ref{fig_3}). The image encoder $E_I$ takes a product image as input, while the text encoder $E_T$ processes the ingredient statement. 
The MLP classifier $F$ combines the image and text embeddings to make the final decision. 

In the image encoder $E_I$, the input image $\mathbf{I}_i$ is initially split into a grid of non-overlapping 32$\times$32 patches, each with 768 channels.
Each patch can be viewed as a token with 768 dimensions. 
These patches are then flattened into vectors and projected into lower-dimensional patch embeddings. 
Additionally, a class embedding and a position embedding are added to each patch embedding. 
These learned embeddings allow the model to incorporate information about the image's classification label and capture the spatial relationships between the patches. 
Layer normalization is applied to normalize feature activations within a single training example.
The patch embeddings are then passed through a series of residual Transformer blocks. 
Each block consists of a 12-head self-attention mechanism and a feedforward neural network. 
The self-attention mechanism enables the model to focus on different parts of the input sequence, while the feedforward neural network learns non-linear relationships between the patch embeddings. 
The final projection layer maps the output of these Transformer blocks to the target output space, enabling predictions to be made for the input images.

The text encoder $E_T$ is also a transformer-based neural network. 
It takes a sequence of text tokens as input, which are initially embedded into fixed-length token embeddings. 
Position embeddings are added to the token embeddings to capture sequential dependencies between the tokens and enhance the representation of the input text. 
The token embeddings are then processed through 12 transformer blocks. 
Each block incorporates a multi-head self-attention layer to capture contextual relationships between the tokens and a feedforward neural network to transform the feature representation. 
Layer normalization is applied after each block to stabilize the training process. 
The final projection layer produces a vector representation of the input text that captures its overall semantic meaning. 

Once the image embeddings from the image encoder and the text embeddings from the text encoder are obtained, the 3-layer MLP classifier $F$ learns a function that maps the combined embeddings to output predictions. 
This classifier combines information from both modalities to make accurate predictions.
\subsection{Training details}
\textit{Pre-training}. We use the pre-trained CLIP officially released by Open AI. 
The pre-training procedure of CLIP was conducted on YFCC100M dataset,\footnote{\url{http://www.multimediacommons.org/}.} a publicly available dataset that contains image-caption pairs, using self-supervised learning, i.e., natural language supervision\cite{radford2021learning}.

\textit{Training.} We provide two solutions to construct the UMDFood model. 
In the first solution, we fix the CLIP model as a feature extractor, and only the MLP classifier was optimized by cross-entropy loss on the UMDFood-90K dataset. 
We further categorize this solution by the input data and refer to the UMDFood model in this solution as UMDFood-VF/LF/LVF. 
Specifically, ``V'' stands for input from the visual modality, i.e., product images, ``L'' stands for input from the language modality, i.e., ingredient statements, and ``F'' is short for features. 
To prepare the input data for training the MLP classifiers, we first feed the ingredient statements and product images into the CLIP model to get the text embeddings and image embeddings, respectively. 
The MLP classifiers of UMDFood-VF, UMDFood-LF, and UMDFood-VLF take the image embeddings, text embeddings, and the concatenation of both image and text embeddings as input, respectively. 
In the second solution, we fine-tune the CLIP model while training the MLP classifier, so both the CLIP model and MLP classifier were optimized by cross-entropy loss on the UMDFood-90k dataset. 
We refer to the UMDFood model in this solution as UMDFood-VL. 
For all solutions, we leverage Adam as the optimizer with a learning rate of 1e-3 for the MLP classifier and 1e-7 for the CLIP model.
The batch size is set to 128, and the number of the training epochs is set to 100. 

\section{Dataset}
\begin{figure*}[!t]
\centering
\includegraphics[width=1\linewidth]{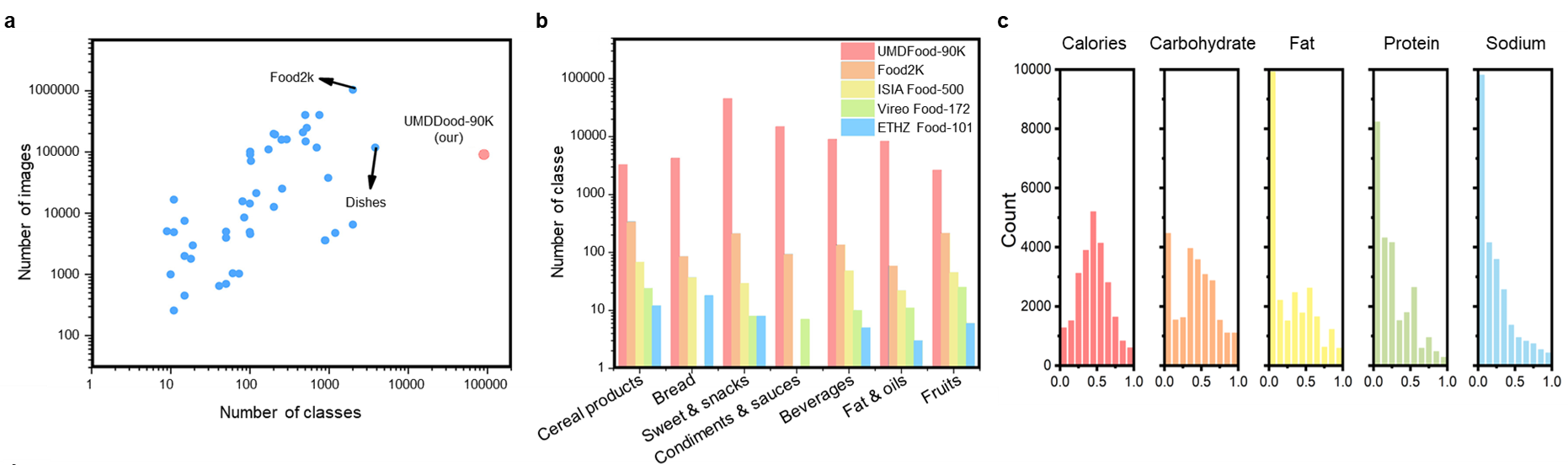}
\caption{The illustration of the UMDFood-90k and UMDFood-VL model. a, comparison of UMDFood-90k with predecessor food databases. UMDFood-90k is at least ten times higher than all other food databases in terms of the diversity of food products collected, i.e., the number of classes in the database. b, the comparison of food categories on UMDFood-90k and typical food database. c, nutrient distribution in UMDFood-90k after normalization. }
\label{fig_4}
\end{figure*}

We hypothesized that the image and ingredient composition can mutually complement with each other to identify the distinctive feature of the food products\cite{jiang2019multi,ma2022neural,liu2021efficient}. We thus first establish the UMDFood-90k dataset, which contained 89,533 food products that covered over 99.9\% of branded food in the US, including 10 broad food groups and 238 categories in the USDA Branded Food Database(fig.~\ref{fig_4}). To date, our database contains more classes and images of food products than any published food database. At the same time, we selected seven major categories to compare with other databases, and UMDFood-90K included more classes than other databases in each category of food products without any significant deviation. Selected nutrients' distribution among UMDFood-90K are also plotted in fig.~\ref{fig_4}(c). The large variation in the distribution of different nutrients leads to the fact that traditional machine learning models often fail to achieve satisfactory results in nutrient estimation tasks. On average, each product contains 12.8 ingredients, and the top 100 high-frequency ingredients have an average document frequency of 12\% (see supporting information). The high sparsity of ingredients poses a challenge for deep-learning models in nutrient estimation tasks.

Our method is trained on the proposed UMDFood-90k dataset, containing 89,533 items. Each item consists of an ingredient statement, and product image, and the values of 5 nutrients including fat, protein, sodium, calories, and carbohydrates. We consider the nutrition value estimation task as a multi-classification task, so we convert the continuous values of each nutrient into discrete classification labels. For each nutrient, we first chose the 0.95 percentile as the threshold and filtered out food products that have higher nutrient values than the threshold to avoid the effect of outliers. Then, we mark the zero value as class 0, sort other non-zero nutrient values in ascending order, and evenly separate them into different classes. To keep the data balanced, the number of classes for these non-zero nutrient values depended on the multiple non-zero nutrient values to zero nutrient values. In the end, there were 4 conditions for fat, 6 conditions for protein, 8 conditions for protein, 38 conditions for calories, and 18 conditions for carbohydrates.

\section{Experiments}
\subsection{Nutrient estimation accuracy on UMDFood-90k}
The UMDFood-VL model consists of a CLIP model and an MLP classifier. The CLIP model takes product images and ingredient statements as input and generates image embeddings and text embeddings to feed into the following 3-layer MLP classifier. During training, we first initialized the CLIP model with the pre-trained weights from Open AI’s CLIP model. Then we optimized the CLIP model and MLP classifier simultaneously using the cross-entropy loss on the UMDFood-90k dataset. 

\begin{figure}
    \centering
    \includegraphics[width=1\linewidth]{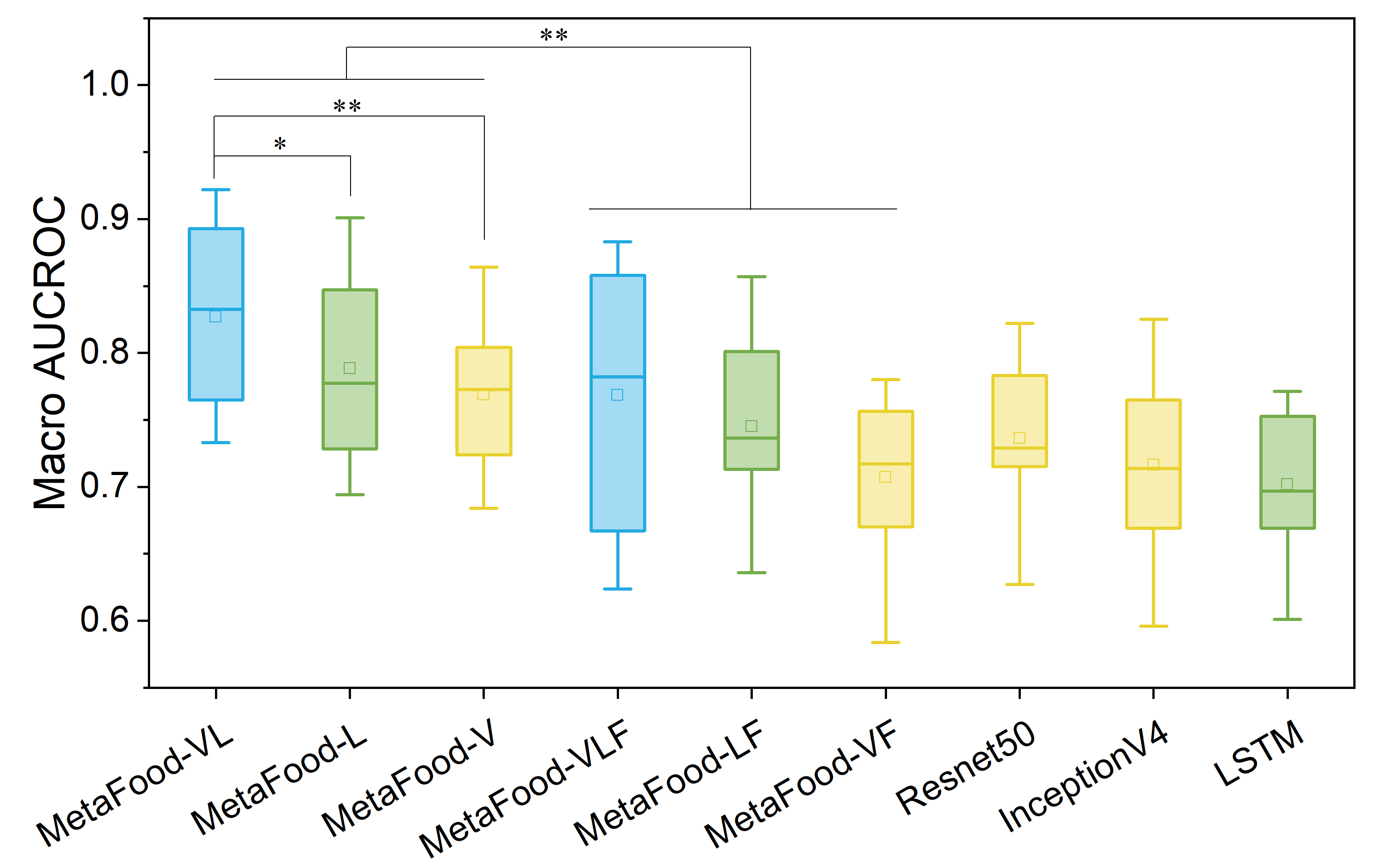}
    \caption{Comparisons of different models on nutrient estimation task based on UMDFood-90k dataset. a, Statistic result of macro-AUCROC of different models on different nutrient (n=11) estimation taskfs.}
    \label{fig_5}
\end{figure}

\begin{figure*}[!t]
    \centering
    \includegraphics[width=1\linewidth]{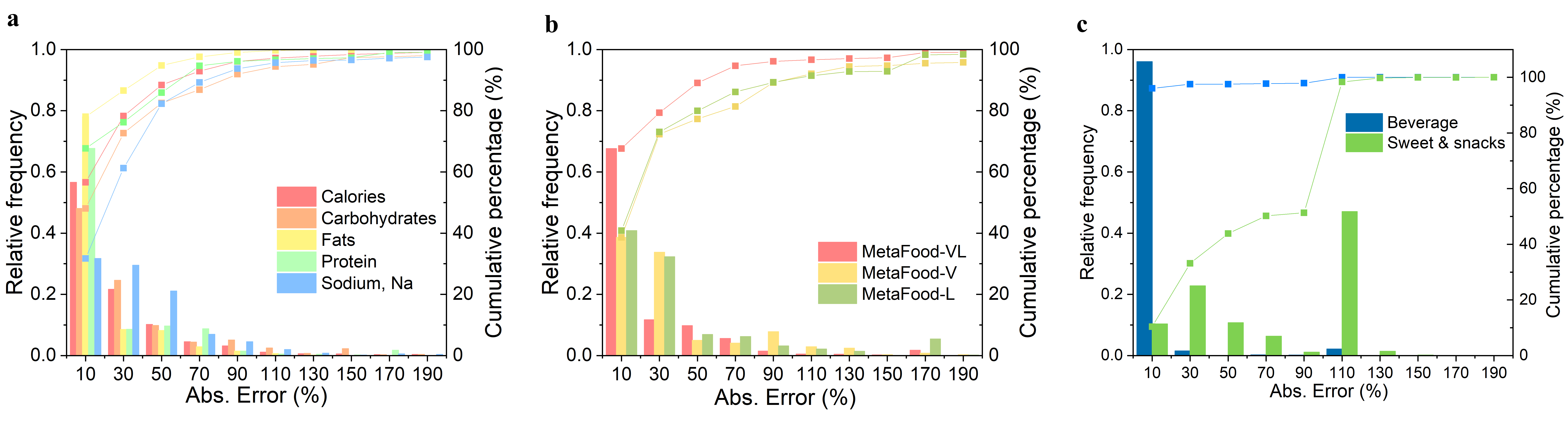}
    \caption{a. The absolute standard error distribution of five selected nutrients by UMDFood-VL model. b. Comparison of absolute standard error distribution of protein content estimation by UMDFood-VL, UMDFood-V, and UMDFood-L models. The multimodal input of image and ingredient information improved the protein estimation task performance when compared to specific input model. c. Absolute standard error distribution of UMDFood-VL model on beverage and sweet \& snack categories calories value in UMDFood-90k database, where UMDFood-VL performed significantly better in beverage. }
    \label{fig_6}
\end{figure*}

We measured the proposed model's performance by macro-AUCROC metrics. Specifically, we split the entire dataset into a training set and a testing set in the 7:3 ratio. We use the testing dataset to calculate the macro-averaging Area Under the Receiver Operating Characteristic Curve (macro-AUCROC) which is commonly used in the evaluation of binary-classification tasks. To adapt this metric to our multi-classification task, we leverage the ``one versus one'' strategy: we consider all possible pairs of different classes, one as the positive class and the other as the negative class. In this way, we can reduce the multi-classification task output into a binary classification one and calculate its AUCROC. We take the unweighted average of AUCROC over all the class pairs as the macro-AUCROC. We also take the weighted average of AUCROC over all the class pairs by the number of true positive instances for each pair. UMDFood-VL outperforms other models significantly ($p <$ 0.08 for UMDFood-L and $p <$ 0.05 for all other models, fig.~\ref{fig_5}. The model achieves an average best nutrient performance of 0.827, with fat estimation reaching 0.921, while calorie estimations are the lowest at 0.709. To aid understanding of the obtained results, we tabulate the frequency distribution of absolute errors measured, with fat value estimation showing 78.0\% and 86.6\% of items among UMDFood-90k absolute error being less than 10\% and 30\%, respectively, satisfying practical accuracy requirements for the food compilation task (fig.~\ref{fig_6}(a)). We compare our model against three state-of-the-art CNN models, including ResNet 50, Inceptive V4, and VGG16, with the ResNet50 achieving the best training result, but still falling 7.2\% short of UMDFood-VL's macro-AUCROC of 0.855 for fat estimation. This demonstrates UMDFood-VL significantly improved the accuracy of the nutrient estimation model when compared to CNN models that rely solely on food images. 

To further demonstrate the benefits and accessibility of multimedia learning, we also trained the CLIP model using only ingredient statements as the input data, named UMDFood-L. The macro-AUCROC of UMDFood-L decreased to 0.788 on average when compared to UMDFood-VL, but 5.2 \% higher than ResNet50. Besides UMDFood-V, we also train CLIP model with sole image data as input, named UMDFood-V. The UMDFood-VL also performs significantly better than UMDFood-V (0.769 on average, $p$ < 0.05). Absolute error distribution validated macro-AUCROC result, where 67.7\% of items' absolute error were less than 10\% for UMDFood-VL model, which decreased to 38.6\% and 40.8\% for UMDFood-L and UMDFood-V, respectively (fig.~\ref{fig_6}(b)).

Furthermore, we evaluated UMDFood-VL's performance in different food categories and found that it performs better in beverage products than in sweet \& snack products for calorie estimation. For instance, 96.0\% of beverage products estimated by the model have an absolute error of less than 10\%, while only 10.4\% of sweet \& snack products met this accuracy requirement (fig.~\ref{fig_6}(c)). More details of model comparison could be found in supporting information.

In conventional CNN models, packaging images of food products contain noise information that interferes with nutrient estimation. For example, images of strawberries may appear in both strawberry-flavored milk and juice, causing the CNN model to focus attention on the strawberry icon and bias the prediction results. Leveraging ingredient statements can recognize more decisive features from the package image and vice versa, leading to more accurate results. Our UMDFood-VL estimation method meets the accuracy and precision requirements of the current food composition compilation task, applying a lenient acceptance range between the database and chemical values for regulations. For example, for nutrients that have a negative impact on health, such as fat, cholesterol, and sodium, the tolerance limit is set at less than 120\%, while nutrients that are good for health, such as vitamins, protein, and dietary fiber, must be higher by 80\% from the nutrition labels. When the calorie data estimate is less than 120\% of the dataset's ground truth value, more than 90.74\% of the products are recognized as accurate. Our work also demonstrates how generative pre-training models can be used to address specific research tasks by using them as encoders and finetuning small-scale neural networks, such as MLP, to achieve numerical estimation.

\subsection{Visualization attention of UMDFood-VL}
\begin{figure*}[!t]
\centering
\includegraphics[width=\linewidth]{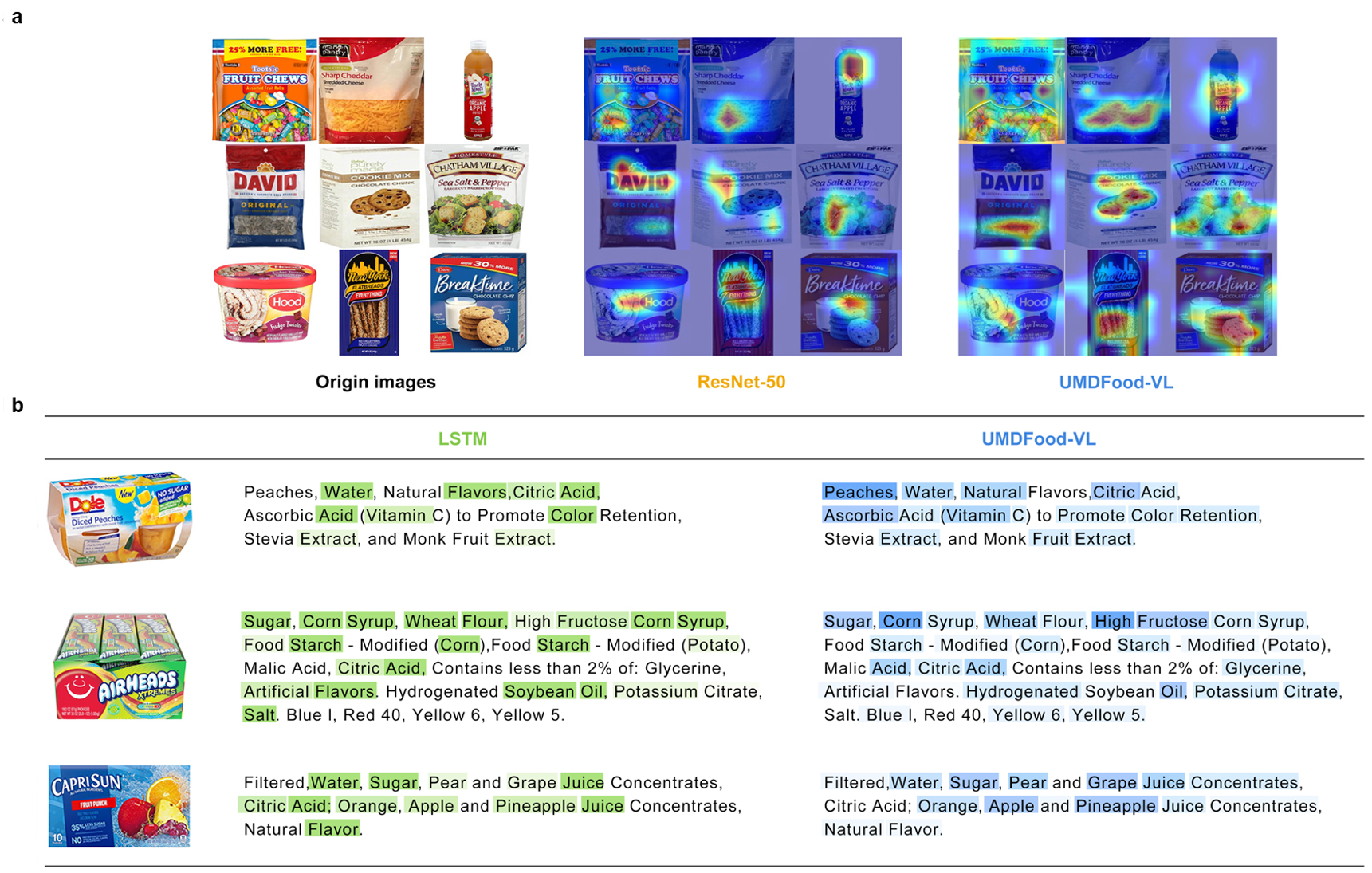}
\caption{Interpretive analysis of different models. a, ResNet50, and UMDFood-VL model attention visualized by GradCAM. b, LSTM and UMDFood-VL model attention on text. A darker color represents a high weight for attention.}
\label{fig_7}
\end{figure*}

To aid in the visual interpretation of our nutrient estimation process from product images, we employ GradCAM\cite{selvaraju2017grad} to produce a coarse localization map highlighting the "attention region" in the food image. Specifically, we use the UMDFood model to generate attention maps based on gradients flowing into the final layer. We find that the presence of ingredients text information caused a significant shift in model attention (fig. ~\ref{fig_7}(a)). Compared to the CNN model, UMDFood-VL appeared to focus more on the food feature regions in the package, such as the cheese and cookie regions in our example images. In contrast, CNN's attention was over-focused, resulting in concentration around a particular feature region and greater susceptibility to diversion. For instance, CNN may focus solely on the croutons in a salad or the brand name in a nut product image. Overall, our attention maps offer a clear visualization of how the model estimates nutrient values from product images and demonstrate the advantages of UMDFood-VL over CNN in terms of attentional focus.

To further elucidate our nutrient estimation process based on ingredient statements, we analyzed the weights assigned to different ingredients by our UMDFood-VL model and compared it to a traditional LSTM model (fig.~\ref{fig_7}(b). We selected three food samples and visualized the weights using a color-coded transparency scheme: higher transparency indicates lower weight, and vice versa. Our results show that the UMDFood-VL model extracted food ingredient features more comprehensively compared to LSTM, resulting in significant differences in the assigned weights of different ingredients. Specifically, LSTM tends to assign higher weights to high-frequency ingredients like "water" and "sugar," while UMDFood-VL assigns higher weights to more specific ingredients such as "peaches" and "apples" in the ingredient statements. These findings demonstrate the impact of multi-media training on the model's ability to accurately estimate nutrient content from ingredient statements. By comprehensively extracting food ingredient features, UMDFood provided more detailed and accurate weight assignments for different ingredients.

\subsection{Model validation by chemical experiments}
\subsubsection{Chemical analysis}
Materials Sodium carbonate (99.5\%), benzoic acid (99.5\%), petroleum ether, potassium chromate (99.5\%), silver nitrate (99.0\%), phenol (98\%), concentrated sulfuric acid (96\%) were all purchased from Sigma-Aldrich (St. Louis, MO, USA). All chemicals are of analytical grade.
Sample preprocessing Weighed 500 g of all beverage samples and then freeze-dried them before subsequent analysis. Record weight after freeze-drying as $W_f$.
Calorie determination by bomb calorimetry. We weighed out 0.5 g of sample, placed it into the holder, and fit it into the pellet press. Then pull down the lever of the pellet press three times and record the value. Calories were measured by Parr 6400 bomb calorimeter which loaded with oxygen automatically and the change in the temperature was measured and logged by the control computer. Benzoic acid was used as the calibration standard with a gross energy of 6.318 kcal/g. 
Total carbohydrate determination by phenol sulfuric acid. We weighed 100 mg of samples into a test tube and heat block at 100 $^{\circ}$C for 3 h with 5 mL of 2.5 M HCl and cooled to room temperature. Then neutralized with 660 mg solid \chemfig{NaCO_3} until the effervescence ceases. Transfer 2.5 mL of sample solution to a centrifuge tube and make up the volume to 50 mL, vortex mixing for 1 min. Subsequently, filtrate the sample solution by filter paper, and collect 2 mL of the final sample solution into a centrifuge tube. Add 5 mL of 96\% sulfuric acid to each tube rapidly and cool down the tubes under a fume hood for 10 min. Read the absorbance at 490 nm for each sample and calculate by standard curve (R$^2$ = 0.9992)\cite{southgate1969determination}.
Protein determination by combustion. First, 1.0 ± 0.2 g of post-freeze-dried samples were weighed in a tin boat. Analyze at least five blanks with tin boats, until three consecutive blanks have a stable value with a standard deviation of less than 0.002\%. Blank correct using the last three consecutive values. Load the set of samples into the protein analyzer. Instrumental settings including the furnace temperature of 1100 $^{\circ}$C, and purge cycles were twice. The baseline delay time is 6 s, the minimum analysis time is 35 s equilibrate time is 30 s, and $T_C$ baseline time is 10 s. The nitrogen factor was set to 6.25\cite{sweeney1989generic}.
Fat determination by Soxhlet method. First, accurately weigh an extraction flask and then add approximately 85 mL of petroleum ether. Then weighted 4.0 g of samples to a thimble. Soxhlet extraction for at least 50 cycles in a minimum of 4 h. Upon completion of extraction, separate the unit and pour off the ether from the Soxhlet extraction apparatus into a large filter positioned on a bottle. Disassemble the Soxhlet apparatus, then put the flask in a steam bath to vaporize the residual petroleum ether until the weight is constant. Fat content of sample $C_\text{Fat}Fat$ was calculated by Eq.~\ref{eq:fat}
\begin{equation}
C_\text{Fat}=0.05(W_a-W_b)W_f
\label{eq:fat}
\end{equation}
where $C_\text{Fat}$ is fat content (g/100g), $W_a$ is the weight of the flask after extraction (g), $W_b$ is the weight of the flask prior to extraction (g), $W_f$ is the weight after freeze-drying \cite{van1949rapid} 

Sodium determination by Mohr titration Pipette accurately 10.00 g of beverage drink samples before freeze-drying in duplicate into 250 mL Erlenmeyer flasks. Added 40 mL of boiling water to each Erlenmeyer flask and stirred the mixture vigorously for 1 min. Then 1 mL of \chemfig{K_2CrO_4} indicator to each 50 mL of prepared sample. Titrate each solution with standardized 0.1 M \chemfig{AgNO_3} (0.1000 ± 0.0005 M) to the first visible pale red-brown color that persists for 30 seconds. Sodium content $C_\text{Sodium,Na}$ of the sample was calculated by Eq.~\ref{eq:Na} 
\begin{equation}
C_\text{Sodium,Na} = 39.07V_\text{Ag$^+$}
\label{eq:Na}
\end{equation}
 $C_\text{Sodium,Na} = 39.07 V_\text{Ag$^+$}$  
where $C_\text{Sodium,Na}$ is content of sodium (g/100g), $V_\text{Ag$^+$}$ is volume of standardized \chemfig{AgNO_3} volume titrant used (mL)\cite{trinder1951rapid}. 
\subsubsection{Model validation by chemical experiment}
\begin{figure*}[!t]
\centering
\includegraphics[width=5in]{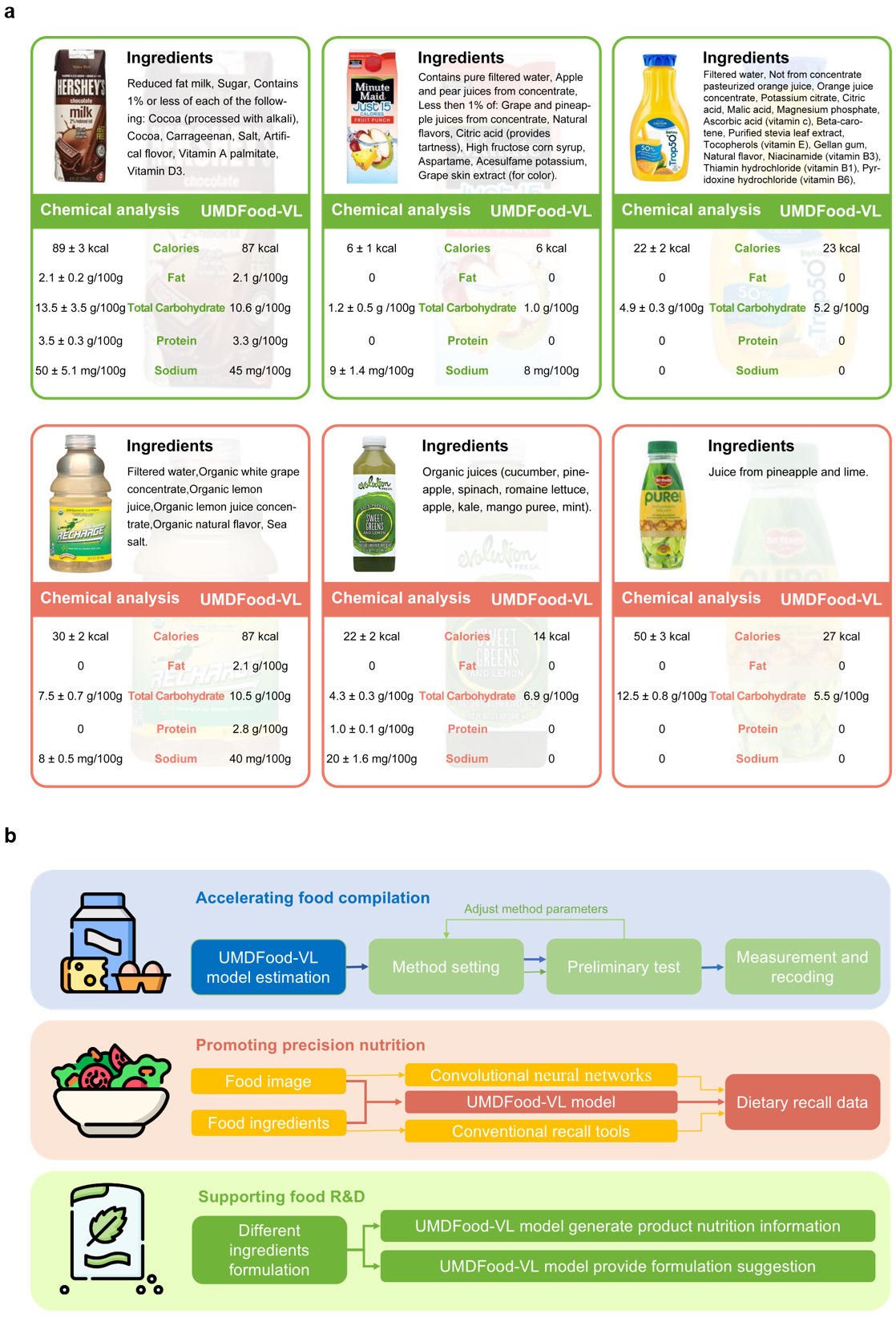}
\caption{The conceptual pipeline of our research. Our contribution includes collecting the UMDFood-90k multimodal dataset and developing UMDFood models for food composition compilation which may boost applications in precision nutrition, food R\&D, and food analysis.}
\label{fig_8}
\end{figure*}

To demonstrate the effectiveness of our nutrient estimation methodology, we conduct direct experimental testing by analyzing 50 randomly selected beverage samples and comparing their chemical analysis values to the values estimated by our model. These samples are chosen to reflect the current demand for new and innovative beverage products, which has made the task of product compilation a challenging one. Overall, our analysis showed that 82.2\% of the selected products estimation error of less than 10\% compared to their chemical analysis values (see in supporting information). To further validate our approach, we compare the estimated values of six representative samples (three successful estimates and three failures) to their corresponding chemical analysis values (fig.~\ref{fig_8}). Our results show that for the samples estimated successfully, the model values fell within the range of the chemical analysis values and database records, confirming that our model met regulatory requirements with high accuracy. However, for the failed samples, the model's estimated values were significantly different from the true values, indicating that these products were unique and did not fit within the general range of comparable products. For example, one such product was a bottle of green yogurt, which was a rare and unconventional product. Another example was pineapple and lime juice, which had a calorie count of 50 kcal/100mL while other juices generally had a range of 20-30 kcal/100mL. These findings highlight the limitations of our methodology and emphasize the importance of carefully selecting representative food samples for accurate nutrient estimation.
In the comparison of the chemical analysis values of the four nutrients and the two data values, we excluded the fat content from the analysis as it was highly concentrated, and most beverage fat contents were 0. We use a 3D scatter plot to visualize the relationships, where the radius of the scatter represents the standard deviation of the chemical analysis results. A more linear scatter distribution indicates a higher correlation between the three relationships.

\begin{figure*}[!t]
\centering
\includegraphics[width=1\linewidth]{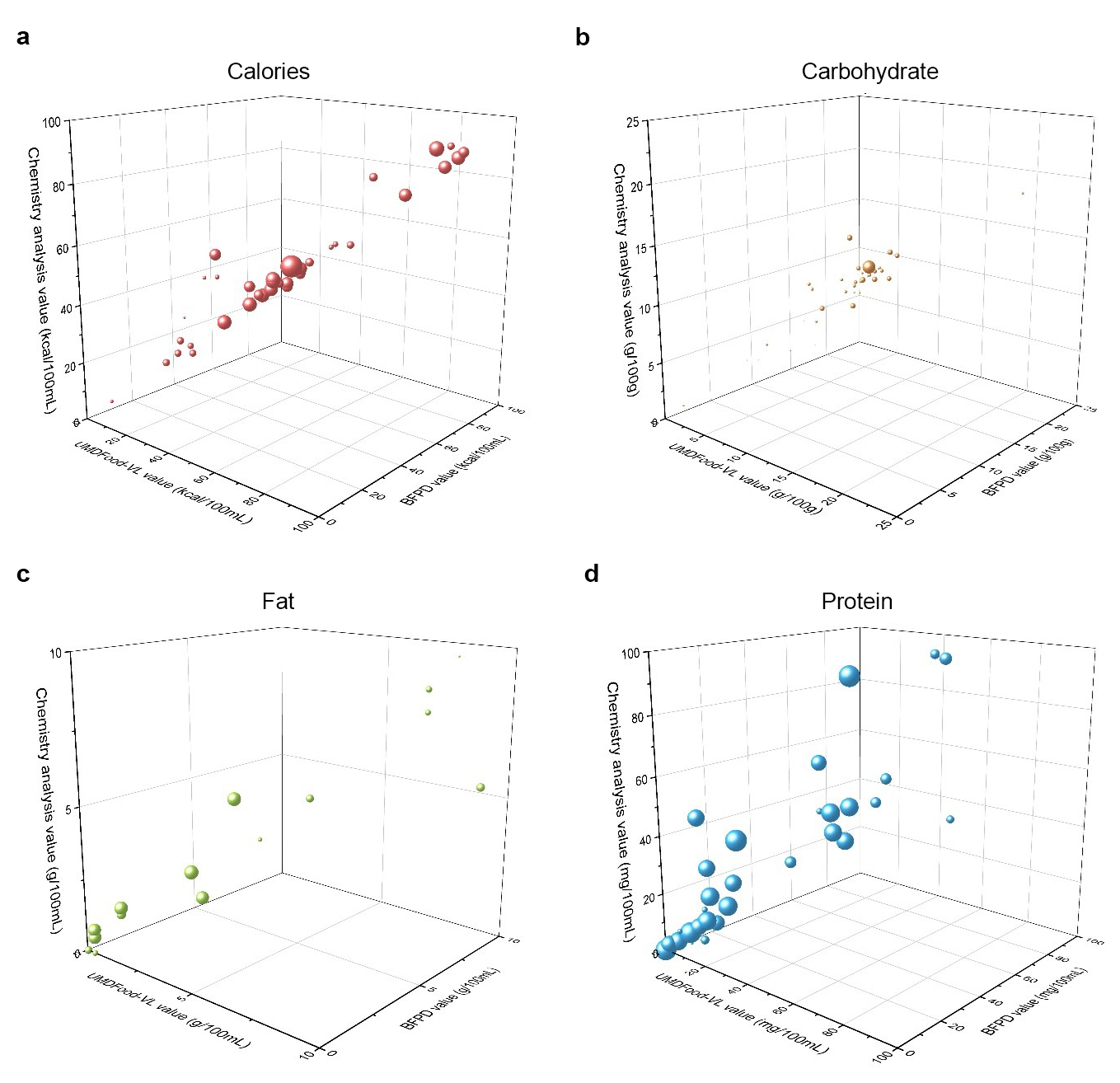}
\caption{Comparison of nutrient value between USDA-BFPD, UMDFood-VL, and chemical analysis value. a, calories, b, carbohydrate by different, c, protein, d, sodium. Three axes represent BFPD value, UMDFood-VL estimated value, and chemistry analysis value, respectively. The radius of each sphere sample is proportional to the chemical analysis error.}
\label{fig_9}
\end{figure*}

The calorie content distribution appears to be relatively evenly distributed. Values with larger deviations tended to have higher chemical analysis values compared to the model-predicted values and BFPD values. This suggests that producers may underestimate the calorie content when labeling their products (fig.~\ref{fig_9}(a)). The distribution of carbohydrate content differed significantly from that of calories. The majority of beverage products have carbohydrate distributions concentrated in the range of 10g/100mL to 15g/100mL, resulting in no significant difference between the BFPD values, UMDFood-VL values, and the chemical analysis values (fig.~\ref{fig_9}(b)). The distribution of protein content is correlated with the content. The figure only includes 15 samples due to the presence of 35 beverage samples with a protein content of 0. Samples with high protein content exhibited larger deviations between the three values. Conversely, for samples with low protein content (less than 5g/100mL), the correlation between the three values is stronger (fig.~\ref{fig_9}(c)). Regarding the estimation of sodium content, the sample distribution is correlated with the content. There are more samples with low sodium content and fewer samples with high sodium content. The error in sodium content testing is higher due to the assay used. For samples with sodium content above 40mg/100mL, the correlation between the three values is poor. Additionally, there is a significant error between the chemical analysis values and the BFPD values used for training. To improve accuracy, it may be beneficial in subsequent optimization to directly test data for products with high sodium content and provide that information to the machine for learning (fig.~\ref{fig_9}(d)).
These observations highlight potential areas for improvement in future iterations of the model and suggest the need for better data collection and testing methods, particularly for nutrients with specific distribution patterns or high variability in the samples.

\section{Discussion}
Compiling information on the nutrient content of conventional foods is a demanding and time-consuming task. To effectively leverage the rich but noisy data available, multimodal modeling, specifically vision-language modeling, is necessary. Our approach employs state-of-the-art vision-language models to simultaneously extract deep information from food images and ingredient statement texts. Joint models provide researchers with a powerful parser and feature extractor that can handle sparse and noisy datasets without requiring tedious text and image preprocessing, while also being robust against data imperfections. Importantly, language models can quantize nutrition representation without needing quantitative information on each ingredient, which is often commercially inaccessible. This is achieved through the accurate alignment of intersected information between images and texts and by synergistically collaborating on complementary information to comprehensively understand food samples. Our UMDFood-VL estimator empowers machines to see, read, comprehend, and analyze multi-media food data, ultimately delivering accurate nutrient estimations.

In addition to meeting the requirements of the food composition compilation task, the UMDFood models also offer new methods for food analysis. By analyzing the weights assigned to different ingredients, we can establish a mapping between ingredients and different nutrients. For instance, in sodium content estimation, not only were food additives containing sodium given high weights but so were "cheese," "wheat," and "yeast." This observation may suggest that bakeries tend to have higher sodium content. Moreover, traditional food chemistry analysis often requires pre-experiments to estimate the nutrient value range of new products. The UMDFood-VF model offers a new estimation method that improves the efficiency of chemical analysis and reduces chemical reagent waste. Additionally, our current personal dietary records are primarily based on either images or text as input. The UMDFood model provides a convenient and accurate solution by allowing users to correct uncertain items in machine image recognition by selecting ingredients. The UMDFood model can also help in the development of new products. For example, we can quantitatively analyze the change in the model-extracted feature vector when replacing color ingredients to identify potential alternatives.

Although the UMDFood model shows promising results, it also has several limitations that need to be taken into account. First, while our self-supervised pre-training approaches based on the CLIP model can increase the labeling efficiency to some extent, our method still requires expert-labeled training data for supervised fine-tuning. As a result, our approach is only able to predict nutrient values that were explicitly annotated in the dataset and cannot predict nutrient values that were not annotated, such as phytochemicals. Second, our AI-empowered solution is data-hungry and relies on the availability of large-scale multi-media food data. However, the creation of such datasets has been a challenging task for researchers, requiring years of effort. Additionally, our model training requires high-performance GPU resources and a significant amount of time (thousands of GPU hours for the pre-trained vision-language model). Third, while we have focused on food images and ingredient texts as our primary multimodal inputs, other modalities such as taste and chemical quantization could further improve our model's performance through joint learning. Fourth, to expand the scope and applicability of our model, we need to enrich the multimodal database with additional food sensory and safety information. This would enable more diverse and practical applications of our model, accelerating the development of precision nutrition and ushering the food industry into a new era. In future research, we aim to explore self-supervised food data learning methods that do not rely on large-scale labeled food data. Such methods would enable us to expand the content of the food multimodal database and enhance the performance of our model.
\section{Conclusion}
In this work, we introduced the UMDFood-90k dataset and the UMDFood-VL model, which used a novel vision-language AI approach for nutrient value estimation. Our model achieved excellent results for fat value estimation, with a macro-AUCROC of up to 0.921. Interpretive analysis showed that our model accurately extracted features from images and ingredient lists, resulting in robust nutrient estimation. We then tested the estimated nutrient values by analyzing 50 randomly selected beverage samples and comparing them to chemical analysis values. The results showed high accuracy, meeting regulatory requirements. Our vision-language model has great potential for food composition compilation and use in precision nutrition and food research.
\section{Acknowledgements}
We acknowledge the financial support of the Maryland Agricultural Experiment Station (MAES) research program (MD-NFSC-232696). We acknowledge the support of Pamela Pehrsson and Jaspreet Ahuja in the U.S. Department of Agriculture Agricultural Research Service (USDA-ARS) Methods and Application of Food Composition Laboratory. We also acknowledge the support of Ying Li from the FDA for food data consulting.

\bibliographystyle{IEEEtran}
\bibliography{UMDFood}

\end{document}